# SODA – a Soft Origami Dynamic utensil for Assisted feeding

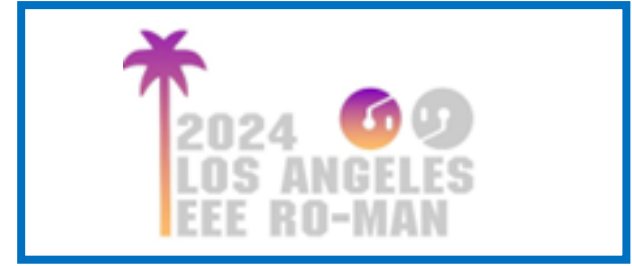

**Yuxin (Ray) Song[1], Shufan Wang[2],**

**[1,2] University of Washington, Seattle, WA**

[1] syx1995@cs.washington.edu
[2] shufaw@uw.edu

SODA aims to revolutionize assistive feeding systems by designing a multi-purpose utensil using origami-inspired artificial muscles. Traditional utensils, such as forks and spoons, are hard and stiff, causing discomfort and fear among users, especially when operated by autonomous robotic arms. Additionally, these systems require frequent utensil changes to handle different food types. Our innovative utensil design addresses these issues by offering a versatile, adaptive solution that can seamlessly transition between gripping and scooping various foods without the need for manual intervention. Utilizing the flexibility and strength of origami-inspired artificial muscles, the utensil ensures safe and comfortable interactions, enhancing user experience and efficiency. This approach not only simplifies the feeding process but also promotes greater independence for individuals with limited mobility, contributing to the advancement of soft robotics in healthcare applications.

## INTRODUCTION

Robot-assisted feeding systems hold significant promise for enhancing the independence and quality of life of individuals with motor impairments. These systems are designed to help users perform the essential task of eating without the continuous need for a caregiver, such as Assistive Dextrous Arm (ADA) [2] and Obi [6]. The development of such systems involves sophisticated technology, including computer vision, haptics, and advanced control algorithms, to ensure precise and safe food handling and transfer.

Various approaches have been explored to optimize the robot-assisted feeding process. Modern systems must address the complex requirements of picking up food, positioning it correctly, and transferring it safely to the user's mouth. Researchers have focused on improving the interaction between the robot and the user, ensuring safety through compliant hardware and control mechanisms, and enhancing user control over the feeding process. These systems are equipped with multiple safety features, such as force thresholds and anomaly detection, to prevent accidents and ensure a comfortable feeding experience.

Traditionally, robot-assisted feeding systems use standard utensils like forks and spoons. While these utensils are familiar and intuitive for human users, they present significant challenges for robotic manipulation. The rigidity and lack of adaptability of traditional utensils can lead to difficulties in securely picking up and transferring food items, often resulting in spills and inefficient feeding.

Related works like the Kiri-Spoon [7] have emerged to address these challenges. The Kiri-Spoon combines the familiar shape of a traditional spoon with a kirigami structure that can adjust its curvature, but it requires a fine fabrication process.

Previous research has applied origami concepts to robot design in several innovative ways, focusing on the adaptability and functionality of origami structures. The magic ball origami structure wheel robot [5] is a notable example that utilizes the magic-ball origami pattern to transform from a long cylindrical tube to a flat circular one, enabling the wheel to adapt to various terrains and reduce mechanical complexity. Another exemplary application is the origami robot for patching stomach wounds, designed to be swallowed and navigate to specific locations within the stomach to perform tasks such as patching wounds [10]. It unfolds from a compact capsule once ingested, demonstrating crucial adaptability for operating in a complex environment [9]. The use of origami in robotics offers increased flexibility, reduces complexity, and leads to lighter, more versatile components that are easier and more cost-effective to manufacture. Additionally, the inherent compliance of origami structures enhances safety and interaction with humans by absorbing impacts and adapting to physical constraints more effectively than rigid components.

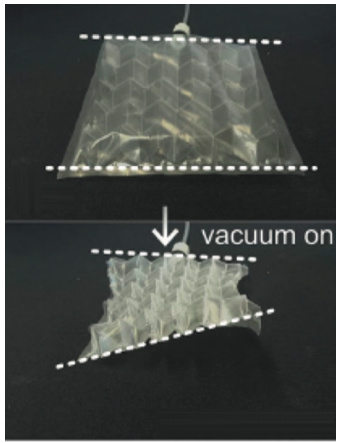

Figure 1. Origami-inspired artificial muscles pattern contracts to achieve shape changing when the vacuum is on [8].

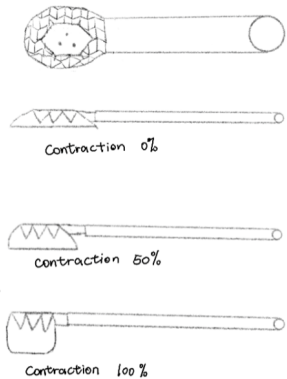

Figure 2. SODA provides a design of single suction cup with different levels of contractions.

## USER RESEARCH

Accessibility in the physical environment remains a significant barrier for many individuals with disabilities [1]. Our target users belong to this accessibility group and face considerable inconveniences during mealtimes, often relying on caregivers for assistance, which can impact their independence and dignity. These users struggle with traditional utensils due to difficulties in gripping, maneuvering, and safely transferring food. These challenges are exacerbated when such utensils are operated by autonomous robotic arms, creating an intimidating and uncomfortable experience. Prior ADA robot systems used hard and sharp forks, which heightened users' fears and discomfort [2].

To address these issues, we conducted extensive user research, focusing on the specific needs and preferences of our target audience. Through interviews, observations, and prototype testing with both able-bodied participants and individuals with disabilities, we gathered crucial insights into the physical and psychological requirements for an effective robot-assisted feeding system. The feedback emphasized the need for a softer, safer, and more intuitive utensil design, capable of handling different types of food without the need for frequent changes. These studies also highlighted the importance of cost-effectiveness and ease of replacement to ensure the system's practicality for everyday use.

## DESIGN PROCESS

The primary objectives for the design of our multi-purpose utensil for robot-assisted feeding include ensuring softness and safety, maximizing efficiency and versatility, maintaining cost-effectiveness, and facilitating easy replacement. Our design aim to avoid hard and sharp edges, making the utensil safe and comfortable for users. Additionally, it aims to combine the functionalities of forking and spooning into a single utensil, ensuring versatility and efficiency in handling various food types. Cost-effectiveness and ease of replacement are also critical considerations, ensuring that the utensil can be produced and replaced affordably and quickly.

A key feature of both designs is the use of fluid-driven origami-inspired artificial muscles [8], the basic pattern is shown in Figure 1. These muscles were chosen for their lightweight, low cost, and high strength, which are essential for creating an effective and practical utensil. Our sketch design (Fig. 2), on the other hand, presents a simplified design inspired by the magic ball structure in origami with a single suction cup that can contract to different levels, providing a straightforward yet effective solution for handling food. This design focuses on minimizing complexity and ensuring ease of use and replacement.

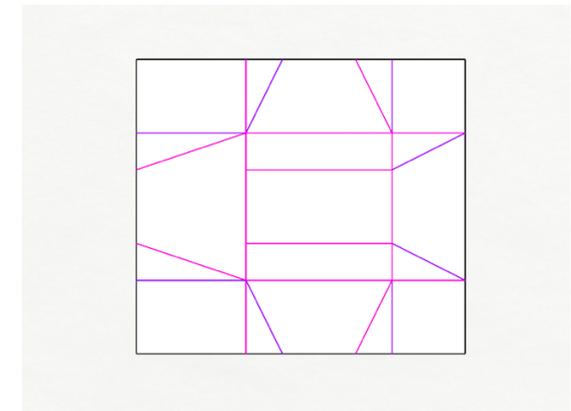

Figure 3. SODA crease pattern

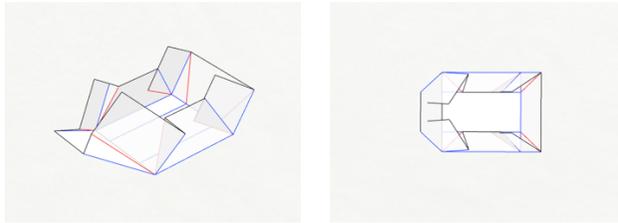

**Figure 4. SODA structure simulation in Rhino 3D, the left simulates 70% contraction for scooping type acquisition, and the right simulates 100% contraction for forking acquisition**.

By inputting and simulating the 2D crease pattern (Fig. 3), the plugin Crane in Grasshopper allows for a visualization (Fig. 4) of deformation in different levels of contraction. The form-finding capabilities of Crane optimize designs under specific constraints, crucial for the creation of utensils that are efficient and straightforward to manufacture. [4]

## CONCLUSION

SODA advances the field of robot-assisted feeding by integrating fluid-driven artificial muscles into a new type of utensils, we achieve a combination of softness, adaptability, and cost-effectiveness, addressing key user needs and preferences. The innovative design leverages the flexibility and simplicity of origami structures, providing a safer and more efficient tool for individuals with motor impairments. This approach not only enhances the interaction between the user and the robot but also simplifies the manufacturing process, making the system more accessible and practical for everyday use. This work underscores the potential of origami principles in revolutionizing assistive devices, offering a promising pathway for future developments in this human-robot interaction.